\RequirePackage{fix-cm}
\documentclass[twocolumn]{svjour3}          % twocolumn
\smartqed  % flush right qed marks, e.g. at end of proof
\usepackage{graphicx}
\usepackage[linesnumbered]{algorithm2e}
\usepackage{amsmath}
\usepackage{amssymb}
\usepackage[bf]{caption}
\usepackage{color}
\usepackage[usenames,dvipsnames]{xcolor} %more colors
\usepackage{listings}
\usepackage{longtable}
\usepackage[super]{nth}
\usepackage{placeins}
\usepackage{siunitx}
\usepackage[font=footnotesize]{subcaption} 
\usepackage{url}
\usepackage{verbatim}

\makeatletter
\newcommand\footnoteref[1]{\protected@xdef\@thefnmark{\ref{#1}}\@footnotemark}
\makeatother

\newcommand{\cols}[2]{{#1(:,1:#2)}}

\newcommand{\CC}{C\nolinebreak\hspace{-.05em}\raisebox{.4ex}{\tiny\bf +}\nolinebreak\hspace{-.10em}\raisebox{.4ex}{\tiny\bf +}}

\DeclareMathOperator*{\rank}{rank}

\journalname{Journal of Mathematical Imaging and Vision}

\begin{document}\sloppy

\title{Background Subtraction using Adaptive Singular Value Decomposition%\thanks{Grants or other notes
}


\author{G\"{u}nther Reitberger         \and
        Tomas Sauer %etc.
}


\institute{G. Reitberger \at
              FORWISS, University of Passau, Passau, Germany \\
              \email{reitberg@forwiss.uni-passau.de}           %  \\
           \and
           T. Sauer \at
              FORWISS, University of Passau, Passau, Germany \\
              \email{sauer@forwiss.uni-passau.de}
}


\maketitle

\begin{abstract}
An important task when processing sensor data is to distinguish
relevant from irrelevant data.  This paper describes a method for an
iterative singular 
value decomposition that maintains a model of the background via
singular vectors spanning a  subspace of the image space, thus
providing a way to determine the amount of new information contained
in an incoming frame. We update the singular vectors spanning the
background space in a computationally efficient manner and provide
the ability to perform block-wise updates, leading
to a fast and robust adaptive SVD computation. The effects of
those two properties and the success of the overall method to perform
a state of the art background subtraction are shown in both
qualitative and quantitative evaluations.

\end{abstract}

\section{Introduction}
\label{sec:intro}

With static cameras, for example in video surveillance, the
background, like houses or trees, stays mostly constant over a series
of frames, whereas the foreground consisting of objects of interest,
e.g. cars or humans, cause differences in image sequences. Background
subtraction aims to distinguish between foreground and background
based on previous image sequences and eliminates the background from
newly incoming frames, leaving only the moving objects contained in
the foreground. These are usually the objects of interest in surveillance.

\subsection{Motivation}
\label{subsec:motivation}
\emph{Data driven} approaches are a major topic in image processing
and computer vision,
leading to state of the art performances, for example in classification or
regression tasks. One example is video surveillance
used for security reasons, traffic regulation, or as information
source in autonomous driving. The main problems
with data driven approaches are that the training data has to be well
balanced and to cover
all scenarios that appear later in the execution phase and has to be
well annotated.
In contrast to cameras mounted at moving objects such as vehicles,
static cameras mounted at some infrastructure observe a
scenery, e.g. houses, trees, parked cars, that is widely fixed or at
least remains static over large amount of frames. If one is interested
in moving objects, as it is the case in the aforementioned
applications, the relevant data is exactly the one different from the
static data. The 
reduction of the input data, i.e., the frames taken from the static
cameras, to the relevant data, i.e., the moving objects, is important
for several applications like the generation of training data for
machine learning approaches or as input for classification tasks
reducing false positive detections due to the removal of the
irrelevant static part.  

Calling the static part \emph{background} and the moving objects
\emph{foreground}, the task of dynamic and static part distinction is
known as foreground background separation or simply \emph{background
  subtraction}.

\subsection{Background Subtraction as Optimization Problem}
\label{subsec:back_sub_as_opti}
Throughout the paper, we make
the assumptions that the camera is static, the background is mostly
constant up to rare changes and illumination, and the moving objects,
considered as foreground, are small relative to the image size. Then
background subtraction can be formulated as an optimization
problem. Given an image sequence stacked in vectorized form into the
matrix $A \in \mathbb{R}^{d\times n}$, with $d$ being the number of
pixels of an image and $n$ being the number of images,
foreground-background separation can be modeled  
as decomposing $A$ into a low-rank matrix $L$, the background
and a sparse matrix $S$, the foreground,
cf.~\cite{Candes:RobustPCA}. This leads to the optimization problem
\begin{equation}
\min_{L,S}\ \rank(L) + \lambda\Vert S\Vert_0 \quad \text{s.t.}
\quad A = L + S. 
\label{eq:low_rank_approx}
\end{equation}
Unfortunately, solving this problem is not feasible. Therefore,
adaptations have to be made. Recall that a singular value
decomposition (SVD) decomposes a matrix $A \in \mathbb{R}^{d\times n}$
into
\begin{equation}
  \label{eq:SVDDef}
  A=U\Sigma V^{T}  
\end{equation}
with orthogonal matrices $U\in \mathbb{R}^{d\times d}$ and $V\in
\mathbb{R}^{n\times n}$ and the diagonal matrix
\[
  \Sigma = \begin{bmatrix}  \Sigma' &\ 0 \\ 0 &\ 0\end{bmatrix} \in
  \mathbb{R}^{d\times n}, \quad \Sigma' \in \mathbb{R}^{r \times r},
  \qquad r = \rank A,
\]
where $\Sigma'$ has strictly positive diagonal values. The SVD makes
no relaxation of the rank, but, given $\ell \le r$, the best (in an
$\ell_2$ sense) 
rank-$\ell$, $\ell\in\mathbb{N}$, estimate $L$ of
$A$ can be obtained by using the first $\ell$ singular values and
vectors, see~\cite{SVD_Approx,SVD_History}. This solves the
optimization problems
\begin{equation}
  \label{EQ:SVD}
  \min \Vert A-L\Vert_F \text{~or~} \Vert A-L\Vert_2 \quad
  \text{s.t.} \quad \rank L \leq \ell.
\end{equation}
We use the following notation throughout our paper: $U_{:,1:\ell} :=
\cols{U}{\ell} := [u_1, ... , u_{\ell}]$, with $u_i$ being the $i$-th
column of U$, i\in\{1, ... , \ell\}$. 

The first $\ell$ columns of the $U$ matrix of the SVD
\eqref{eq:SVDDef} of $A$, i.e., the left singular vectors corresponding
to the $\ell$ biggest singular values, span a subspace of the column
space of $A$. The background of an image $J \in \mathbb{R}^{d\times
  1}$ is calculated by the orthogonal projection of $J$ on $U_\ell :=
U_{:,1:\ell}$ by $U_\ell (U_\ell^T J)$. The foreground then consists of the
difference of the background from the image $J - U_\ell (U_\ell^T
J) = \left( I - U_\ell U_\ell^T \right) J$. 

The aim of a surveillance application is to subtract the background
from every incoming image. Modeling the background via \eqref{EQ:SVD}
results in a batch algorithm, where the low rank approximations are
calculated based on some (recent) sample frames stacked together to
the matrix $A$. Note that this allows the background to change slowly
over time, for example due changing illumination or to parked cars
leaving the scene. It is well--known that
the computational effort to determine the SVD of $A$ with dimensions
$d \gg n$ is $O(d n^2)$ using R-SVD and computing only $U_n =
U_{:,1:n}$ instead of the complete $d\times d$ matrix $U$, and the
memory consumption is $O(d n)$, cf.~\cite{Golub_van_Loan}. Especially
in the case of higher definition images, only rather few samples $n$
can be used in this way. This results in a dependency of the
background model from the sample image size and an inability of
adaption to a change in the background that is not covered in the few
sample frames. Hence, a naive batch algorithm is not a suitable solution.

\subsection{Main Contributions and Outline}
\label{subsec:main_contrib}
The layout of this paper is as follows. In Sec.~\ref{sec:rel_work} we
briefly revise related work in background subtraction and SVD
methods. Sec.~\ref{sec:SVD_update} introduces our algorithm of
iteratively calculating a SVD.  The main contribution here consists in
the application and adaption of the iterative SVD to background
subtraction. In Sec.~\ref{sec:algorithm} we propose a concrete
algorithm that adapts the model of the background in a way that is
dependent on the incoming data because of which we call it
\emph{adaptive SVD}. A straightforward version of the algorithm still
has limitations, because of which we present extensions of the basic
algorithm that overcome these deficits.
In Sec.~\ref{sec:comp_results} evaluations of the method give an impression on execution time, generality, and performance capabilities of the adaptive SVD. Finally, in Sec.~\ref{sec:conclusions} our main conclusions are outlined.

\section{Related Work}
\label{sec:rel_work}
The ``philosophical'' goal of background modeling is to acquire a
background image that does not include any moving objects. In
realistical environments, the background may also change, due to
influences like illumination or objects being introduced to or removed
from the scene.
Taking into account these problems as well as robustness and
adaptation, background modeling methods can, according to the survey
papers~\cite{BOUWMANS_1,BOUWMANS_2,BOUWMANS_3}, be classified into the
following categories: Statistical Background Modeling, Background Modeling via Clustering, Background Estimation and Neural Networks.

The most recent approach is, of course, to model the background via
neural networks. Especially convolutional neural networks
(CNNs)~\cite{CNN} have performed very well in may tasks of image
processing. These techniques, however, usually involve a labeling of
the data, i.e., the background has to be annotated, mostly manually, for a set
of \emph{training images}. The network then learns the background
based on the labels. Background modeling is often combined with
classification or segmentation tasks where every pixel of an image is
assigned to one class. Based on the classes, the pixel can then be
classified as background or foreground, respectively. Such techniques
strongly depend on the trained data and besides new approaches like
transfer learning~\cite[p.~526]{DeepLearning} or reinforcement
learning~\cite{ReinforcementLearning} can only be improved by adding
new data.

Statistical background modeling includes Gaussian models, support
vector machines and subspace learning models. Subspace learning
originates from the modeling of the background subtraction task as
shown in~\eqref{eq:low_rank_approx}. Our approach therefore also
belongs to this domain. Principal Component Pursuit
(PCP)~\cite{Candes:RobustPCA} is based on the convex relaxation
of~\eqref{eq:low_rank_approx} by 
\begin{equation}
\label{EQ:PCP}
\min_{L,S} \Vert L \Vert_* + \lambda\Vert S\Vert_1 \quad \text{s.t.} \quad A = L + S,
\end{equation}
with $\Vert L \Vert_*$ being the nuclear norm of matrix $L$, the sum
of the singular values of $L$. 
The relaxation \eqref{EQ:PCP} can be solved by efficient algorithms
such as alternating optimization. As PCP considers the $\ell_1$ error,
it is more robust against outliers or salt and pepper noise than SVD
based methods and thus more suited to situations that suffer of that
type of noise. Since outliers are not a substantial problem in traffic
surveillance which is our main application in mind, we do not have to
dwell on this type of robustness. In addition, the pure PCP method
also has its limitations such as being a batch algorithm, being
computationally expensive compared to SVD, and maintaining the exact
rank of the low rank approximation, cf.~\cite{PCP_overview}. This is a
problem when it comes to data that is affected by noise in most
components, which is usually the case in camera based image
processing. We remark that to overcome the drawbacks of plain PCP,
many extensions of the PCP have been introduced,
see~\cite{PCP_overview,IncPCP_paper}. 

There is naturally a close relationship between our SVD based approach
and incremental principal component analysis (PCA) due to the close
relationship between SVD and PCA. Given a matrix $A \in
\mathbb{R}^{n\times d}$ with $n$ being the number of samples and $d$
the number of features, the PCA searches for the first $k$
eigenvectors of the correlation matrix $A^T A$ which span the same
subspace as the first $k$ columns of the $U$ matrix of the SVD of
$A^T$, i.e., the left singular vectors of $A^T$. Thus, usually the PCA
is actually calculated by a SVD, since $A^T = U\Sigma V^T$ gives $A^T
A = U\Sigma V^T V \Sigma U^T = U\Sigma^2 U^T$, the PCA produces the
same subspace as our iterative SVD approach. One difference is that
PCA originates from the statistics domain and the applications search
for the main directions in which the data differs from the mean data
sample. That is why the matrix $A$ usually gets normalized by
subtraction of the columnwise mean and divided by the columnwise
standard deviation before calculating the PCA which, however, makes no
sense in our application. This is also expressed in the work by Ross
et al.~\cite{IncPCA}, based on the sequential Karhunen-Loeve basis
extraction from~\cite{Seq_Kahrunen_Loeve}. They use the PCA as a
\emph{feature extractor} for a tracking application. In our approach,
we model the mean data, the background, by singular vectors only and
dig deeper into the application to background subtraction, which we
have not seen in works has not been considered in the PCA
context. Nevertheless, we will make further
comparisons to the PCA approach, pointing out further similarities and
differences to our approach.

\section{Update methods for rank revealing decompositions and
  applications} 
\label{sec:SVD_update}

Our background subtraction method is based on an iterative calculation
of an SVD for matrices augmented by columns, cf.~\cite{Sauer19}. In
this section we revise the essential statements and the advantages of
using this method for calculating the SVD.

\subsection{Iterative SVD}
\label{subsec:IterSVD}
The method from \cite{Sauer19} is outlined, in its basic form,
as follows:
\begin{itemize}
\item Given: SVD of $\mathbb{R}^{d\times n_k} \ni A_k =
  U_k\Sigma_k V_k^{T}, n_k \ll d$, and $\rank(A_k) =: r_k$ 
\item Aim: Compute SVD for $A_{k+1} = [A_k, B_k],$\\ $B_k \in
  \mathbb{R}^{d\times m_k},\ m_k := n_{k+1} - n_k$ 
\item Update: $A_{k+1} = U_{k+1}\Sigma_{k+1} V_{k+1}^T$ with\\
  \[
    U_{k+1} = U_k Q \begin{bmatrix} \tilde{U} & 0 \\ 0 &
      I \end{bmatrix},
  \] 
  \[
    V_{k+1} = \begin{bmatrix} V_k & 0 \\ 0 & I \end{bmatrix}
    (P'_k P_k)^T \begin{bmatrix} \tilde{V} & 0 \\ 0 &
      I \end{bmatrix},
  \]
  where $Q$ results from a QR-decomposition, $\Sigma_{k+1}$,
  $\tilde{U}$ and $\tilde{V}$ result from the SVD of a $(r_k+m_k)
  \times (r_k+m_k)$ matrix. $P_k$ and $P'_k$ are permutation
  matrices. 
\end{itemize}
For details, see \cite{Sauer19}.
In the original version of the iterative SVD, the matrix $U_k$ is
(formally) of dimension $d\times d$. Since in image processing $d$
captures the amount of pixels of one image, an explicit representation
of $U_k$ consumes too much memory to be efficient which suggests to
represent $U_k$ in terms of Householder reflections. This ensures that
the \textit{memory consumption} of the SVD of $A_k$ is bounded by
$O(n_k^2 + r_k d)$,
and the step $k+1$ requires
$O(n_{k+1}^3 + d\, m_k(r_k + m_k))$
\textit{floating point operations}.

\subsection{Thresholding - Adaptive SVD}
\label{subsec:thresholding}
There already exist iterative methods to calculate an SVD, but for our
purpose the approach from \cite{Sauer19} has two favorable
aspects. The first one is the possibility to perform blockwise updates
with $m_k > 1$, that is, with several frames. The second one is the
ability to estimate the effect of appending $B_k$ on the singular
values of $A_{k+1}$. In order to compute the SVD of $A_{k+1}$, $Z :=
U_k^T B_k$ is first calculated and a QR decomposition with column
pivoting of $Z_{r_k+1:d,:} = Q R P$ is determined. The $R$ matrix
contains the information in the added data $B_k$ that is not already
described by the singular vectors in $U_k$. Then, the matrix $R$ can
be truncated by a significance level $\tau$ such that the singular
values less than $\tau$ are set to zero in the SVD calculation of
\[
  \begin{bmatrix} \Sigma_k' & Z_{1:r_k,:}P^T \\&
    R\end{bmatrix}.
\]
Therefore, one can determine only from the (cheap)
calculation of a QR decomposition, whether the new data contains
significant new information and the threshold level $\tau$ can control
how big the gain has to be for a data vector to be added to the
current SVD decomposition in an iterative step.

\section{Description of the Algorithm}
\label{sec:algorithm}
In this section, we give a detailed description of our algorithm to
compute a background separation based on the adaptive SVD.

\subsection{Essential Functionalities}
\label{subsec:functionalities}
The algorithm in~\cite{Sauer19} was initially designed with the
goal to determine the kernels of a sequence of columnwise augmented
matrices using the $V$ matrix of the SVDs. In
background subtraction, on the other hand, we are interested in finding a
low rank approximation of the column space of $A$ and therefore
concentrate on the $U$ matrix of the SVD which will us allow to avoid
computation and storage of $V$.
 
The adaptive SVD algorithm starts with an initialization step called
\textit{SVDComp}, calculating left singular vectors and singular
values on an initial set of data. Afterwards, data is added
iteratively by blocks of arbitrary size. For every frame in such a
block, the foreground is determined and then the \textit{SVDAppend}
step performs a thresholding described in
Sec.~\ref{subsec:thresholding} to check whether the frame is
considered in the update of the singular vectors and values that
correspond to the background. 

\subsubsection{SVDComp}
\label{subsec:SVDComp}
\textit{SVDComp} performs the initialization of the iterative
algorithm.
It is given the matrix
$A \in \mathbb{R}^{d\times n}$ and a column number $\ell$ and computes
the best rank-$\ell$ approximation $A = U\Sigma V^T$,
\begin{align*}
U =:[U_{0}, U_{0}']\text{, }\Sigma =: \begin{bmatrix}\Sigma_0 & 0 \\ 0 & \Sigma_0'\end{bmatrix}\text{, }V =:[V_0, V_0']\text{,}
\end{align*}
by means of an SVD
with $\Sigma_0 \in \mathbb{R}^{\ell\times \ell}, U_0 \in
\mathbb{R}^{d\times \ell}$, and $V_0 \in \mathbb{R}^{n\times
  \ell}$. Also this SVD is conveniently computed by means of the
algorithm from \cite{Sauer19}, as the thresholding of the augmented
SVD will only compute and store an at most rank-$\ell$ approximation,
truncating the $R$ matrix in the augmentation step to at most $\ell$
columns. This holds both for initialization and update in the
iterative SVD. 

As mentioned already in Sec.~\ref{subsec:IterSVD}, $U_0$ is not stored
explicitly but in the form of Householder vectors $h_j, j =
1,\dots,\ell$, stored in a matrix $H_0$. Together with a small matrix
$\widetilde{U}_0 \in \mathbb{R}^{\ell \times \ell}$ we then have
\[
  U_0 = \widetilde{U}_0 \prod_{j=1}^{\ell} (I - h_j h_j^T),
\]
and multiplication with $U_0$ is easily performed by doing $\ell$
Householder reflection and then multiplication with an $\ell \times
\ell$ matrix.
Since $V_0$ is not needed in the algorithm it is neither computed nor
stored.

\subsubsection{SVDAppend}
\label{subsubsec:svdappend}
This core functionality augments a matrix $A_k$, given by
$\widetilde{U}_k, \Sigma_k, H_k$, determined either by
\textit{SVDComp} or previous applications of \textit{SVDAppend}, by
$m$ new frames contained in the matrix $B \in \mathbb{R}^{d\times
  m}$ as described in Sec.~\ref{subsec:IterSVD}. The details of this
algorithm based on Householder representation can be found
in~\cite{Sauer19}. By the thresholding procedure from
Sec.~\ref{subsec:thresholding} one can determine, even before the
calculation of the SVD, if an added column is significant relative to
the threshold level $\tau$. This saves computational capacities by
avoiding the expensive computation of the SVD for images that do not
significantly change the singular vectors representing the
background.

The choice of $\tau$ is significant for the performance of the
algorithm. The basic assumption for the adaptive SVD is that the
foreground consists of \emph{small} changes between
frames. Calculating \textit{SVDComp} on an initial set of frames and
considering the singular vectors, i.e., the columns of $U_0$, and the
respective singular values gives an estimate for the size of the
singular values that correspond to singular vectors describing the
background. With a priori knowledge of the maximal size of foreground
effects, $\tau$ can even be set absolutely to the size of singular
values that should be accepted. Of course, this approach requires
domain knowledge and is not entirely data driven.

Another heuristic choice of $\tau$ can be made by considering the
difference between two neighboring singular values $\sigma_i -
\sigma_{i+1}$, i.e., the discrete slope of the singular values. The
last and smallest singular values describe the least dominant
effects. These model foreground effects or small effects, negligible
effects in the background.
%The difference of neighboring singular values is small.
With increasing singular values, the importance of the singular
vectors is growing. Based on that intuition, one can set a threshold
for the difference of two consecutive singular values and take the
first singular value exceeding the difference threshold as
$\tau$. Fig.~\ref{fig:stock_effect_4} illustrates a typical
distribution of singular values. Since we want the method to be
entirely data driven, we choose this approach. The threshold $\tau$ is
determined by $\hat{i} := \min\left\{i:\sigma_i - \sigma_{i+1} < 
\tau^*\right\}$ and $\tau=\sigma_{\hat{i}}$ with the threshold
$\tau^*$ of the slope being determined in the following.

\subsubsection{Re-initialization}
\label{subsubsec:reinitialization}
The memory footprint at the $k$-th step in the algorithm described in
Sec.~\ref{subsec:IterSVD} is $O(n_k^2 + r_k\, d)$ and
grows with every frame added in the
\textit{SVDAppend} step. Therefore, a re-initialization of the decomposition
is necessary.    

One possibility is to compute an approximation of $A_k \approx U_k
\Sigma_k V_k^T \in \mathbb{R}^{d\times n_k}$ or the exact matrix $A_k$
by applying \textit{SVDComp} to $A_k$ with a rank limit of $\ell$ that
determines the number of singular vectors after re-initialization. This
strategy has two disadvantages.
The first one is that this needs $V_k$, which is otherwise not needed
for modeling the background, hence would require unnecessary
computations. Even worse, though
$\widetilde{U}_0 \in \mathbb{R}^{\ell\times \ell}$,
$\Sigma_0 \in \mathbb{R}^{\ell\times \ell}$, and $H_0 \in
\mathbb{R}^{d\times \ell}$ are reduced properly, the memory
consumption of $V_0 \in \mathbb{R}^{n_k\times \ell}$ still depends on
the number of frames added so far. 

The second re-initialization strategy, referred to as (II), builds on
the idea of a rank-$\ell$ approximation of a set of frames
representing mostly the background. For every frame $B_i$ added in
step $k$ of the \textit{SVDAppend} the orthogonal projection
\[
  \cols{U_k}{\hat{i}} (\cols{U_k}{\hat{i}}^T B_i),
\]
i.e. the ``background part'' of $B_i$, gets stored successively. The
value $\sigma_{\hat{i}}$ is determined in
Sec.~\ref{subsubsec:svdappend} as threshold for the \textit{SVDAppend}
step. If the number of stored background images exceeds a fixed size
$\mu$, the re-initialization gets performed via \textit{SVDComp} on
the background images. No matrix $V$ is necessary for this strategy
and the re-initialization is based on the background projection of the
most recently appended frames.

In the final algorithm we use a third strategy, referred to as (III)
which is inspired by the sequential Karhunen-Loeve basis
extraction~\cite{Seq_Kahrunen_Loeve}. The setting is very similar and
the $V$ matrix gets dropped after the initialization as well. The
update step with a data matrix $B_k$ is performed just like the update
step of the iterative SVD calculation in Sec.~\ref{subsec:IterSVD}
based on the matrix $[U_k\Sigma_k, B_k]$. The matrices $\Sigma_{k+1}$
and $U_{k+1}$ get truncated by a thresholding of the singular values
at every update step. Due to this thresholding, the number of singular
values and accordingly the number of columns of $U_k$ has an upper
bound. Therefore, the maximum size of the system is fixed and no
re-initialization is necessary. Calculating the SVD of $[U_k\Sigma_k,
B_k]$ is sufficient since due to 
\begin{align*}
  [U_k\Sigma_k, B_k] &[U_k\Sigma_k, B_k]^T = U_k\Sigma_k \Sigma_k^T
                       U_k^T + B_k B_k^T\\ 
                     &= U_k\Sigma_k V_k^T V_k \Sigma_k^T U_k^T + B_k B_k^T\\
                     &= [U_k\Sigma_k V_k^T, B_k] [U_k\Sigma_k V_k^T,
                       B_k]^T
\end{align*} 
the eigenvectors and eigenvalues of the correlation matrices with
respect to $[U_k\Sigma_k, B_k]$ and $[U_k\Sigma_k V_k^T, B_k]$
are the same. Therefore, the the singular
values of $[U_k\Sigma_k, B_k]$ and 
$[U_k\Sigma_k V_k^T, B_k]$ are the same, being roots of the
eigenvalues of the correlation matrix. In our approach we combine the
adaptive SVD with the re-initialization based on $U_k \Sigma_k$,
i.e. we perform \textit{SVDComp} on $U_k \Sigma_k$, because we want to
keep the thresholding of the adaptive SVD. This is essentially the
same as an update step in Karhunen-Loeve setting with $B_k = 0$ and a
more rigorous thresholding or a simple truncation of $U_k$ and
$\Sigma_k$.
The thresholding strategy of the adaptive SVD
Sec.~\ref{subsec:thresholding} is still valid, as the QR-decomposition
with column pivoting sorts the columns of the matrix according to the
$\ell_2$ norm and the columns of $U_k\Sigma_k$ are ordered by the
singular values due to $||U \Sigma_{:,i}||_2 = \sigma_i$. $U_k
\Sigma_k$ already is in SVD form and therefore \textit{SVDComp} at
re-initialization is reduced to a QR decomposition to regain
Householder vectors and a truncation of $U_k$ and $\Sigma_k$ which is
less costly than performing a full SVD.

Since it requires the $V$ matrix, the first re-initialization strategy
will not be considered in the following, where we will compare only the
strategies (II) and (III). 

\subsubsection{Normalization}
\label{subsubsec:normalization}

The concept of re-initialization via a truncation of $U_k$ and
$\Sigma_k$ either directly through \textit{SVDComp} of $U_k \Sigma_k$
or in the Karhunen-Loeve setting with thresholding of the singular
values still has a flaw: the absolute value of the singular values
grows with each frame appended to $U_k \Sigma_k$ as
\[
  \sum_{i=1}^{n}\sigma_i^2 = \| A \|_F^2.
\]
This also accounts for
\begin{align*}
  \sum_{i=1}^{n_{k+1}}\sigma_{n_{k+1},i}^2
  &= \| U_{k+1} \Sigma_{k+1} \|_F^2
    \approx \left\| [U_k \Sigma_k, B_k] \right\|_F^2 \\
  &= \| U_k \Sigma_k \|_F^2 + \| B_k \|_F^2. 
\end{align*}
The approximation results from the thresholding performed at the
update step. As only small singular values get truncated, the sum of
the squared singular values grows essentially with the Frobenius norm
of the appended frames. Growing singular values do not only introduce
numerical problems, they also deteriorate thresholding strategies and
the influence of newly added single frames decreases in later steps of
the method. Therefore, some upper bound or normalization of the
singular values is necessary. 

Karhunen-Loeve~\cite{Seq_Kahrunen_Loeve} introduce a \emph{forgetting 
factor} $\varphi \in [0,1]$ and update as $[\varphi \, U_k\Sigma_k,
B_k]$. They 
motivate this factor semantically: more recent frames get a higher
weight. Ross et al.~\cite{IncPCA} show that this value limits the
observation history. With an appending block size of $m$ the effective
number of observations is $m/(1-\varphi)$. By the Frobenius norm argument,
the singular values then have an upper bound. By the same motivation,
the forgetting factor could also be integrated into strategy
(III). Moreover, due to
\[
  \| (\varphi \, U_k \Sigma_k)_{:,i}\|_2
  = \left\| \varphi \, \sigma_i \, U_{:,i} \right\|_2 = \varphi
  \sigma_i,
\]
the multiplication with the forgetting factor keeps the order of the
columns of $U_k\Sigma_k$ and linearly affects the 2-Norm and is thus
compliant with the thresholding. However, the concrete choice of the
forgetting factor in unclear.

Another idea for normalization is to set an explicit upper bound for
the Frobenius norm of observations contributing to the iterative SVD,
or, equivalently, to $\sum \sigma_i^2 = \|A\|_F^2$. At initialization,
i.e. at the first \textit{SVDComp}, the upper bound is determined by
$\frac{\|A\|_F^2}{n}\eta$ with $n$ being the number of columns of $A$
and $\eta$ being the predefined maximum size of the system. This upper
bound is a multiple of the mean squared Frobenius norm of an input
frame and we define a threshold $\rho :=
\frac{\|A\|_F}{\sqrt{n}}\sqrt{\eta}$. If the Frobenius norm
$\|\Sigma_0\|_F$ of the singular values exceeds $\rho$ after a
re-initialization step, $\Sigma_0$ gets normalized to $\Sigma_0
\frac{\rho}{\|\Sigma_0\|_F}$. One advantage of this approach is that
the effective system size can be transparently determined by the
parameter $\eta$.

In data science, normalization usually aims for zero mean and standard
deviation one. Zero mean over the pixels in the frames, however leads
to subtracting the row wise mean of $A$, replacing $A$ by $(I - 1 1^T)
A$. This approach is discussed in incremental PCA, cf.~\cite{IncPCA},
but since the mean image usually contributes substantially to the
background, it is not suitable in our application.

A framewise unit standard deviation makes sense since the standard
deviation approximates the contrast in image processing and we are
interested in the image content regardless of the often varying
contrast of the individual frames. Different contrasts on a zero mean
image can be seen as a scalar multiplication which also applies for
the singular values.
Singular values differing with respect to the contrast are not a
desirable effect which is compensated by subtracting the mean and
dividing by the standard deviation of incoming frames $B$, yielding
$\frac{B-\mu}{\sigma}$.
Due to the normalization of single images, the upper bound for the
Frobenius norm $\rho$ is more a multiple of the Frobenius norm of an
average image.

\subsection{Adaptive SVD Algorithm}
\label{subsec:adaptiveSVD}
The essential components being described, we can now sketch our method
based on the adaptive SVD in Alg.~\ref{alg:AdaptiveSVD}.

%\vspace{-0.5em}
\begin{algorithm}[htb]
	\footnotesize
	%\caption{Background Subtraction using adaptive SVD.}
	
	\SetKwFunction{SVDComp}{SVDComp}
	\SetKwFunction{SVDAppend}{SVDAppend}
	\SetKwData{SVD}{SVD}
	\SetKwData{J}{J}
	\SetKwData{B}{B}
	\SetKwData{F}{F}
	\SetKwData{S}{S}
	
	\KwData{Images of a static camera and a matrix $A$ of initialization images.}
	\KwResult{Background and foreground images for every input image.}
	
	$U, \Sigma, \hat{i}$ $\leftarrow$ {\color{blue}\SVDComp{A, $\ell$, $\tau^*$}}\;
	\vspace{0.3em}
	\While{there are input images}{
		\B $\leftarrow$ read, vectorize, and normalize the current image\;	
		
		{\color{gray}\emph{// project B onto the current background model}\;}
		\J $\leftarrow$ $U_{:,1:\hat{i}} (U_{:,1:\hat{i}} ^T \B)$\; 
		
		{\color{gray}\emph{// subtract the background from B and use this as mask on the input image}\;}
		{\color{olive}\F} $\leftarrow$ $\B\, \cdot (|\B - \J| > \theta)$\; 
		
		{\color{gray}\emph{// build a block of input images}\;}
		$M \leftarrow [M, \B]$\;  
		
		{\color{gray}\emph{// append a block of images}\;}
		\If{$M.cols == \beta$}{
			
			$U, \Sigma, \hat{i}$ $\leftarrow$ {\color{blue}\SVDAppend{$U$, $\Sigma$, M, $\tau^*$}}\;
			$M \leftarrow $ [ ]\;
		}
		
		{\color{gray}\emph{// re-initialization if maximum size is exceeded}\;}
		\If{$U.cols > n^*$}{
			
			$U, \Sigma$ $\leftarrow$ {\color{blue}\SVDComp{$U\Sigma$, $\ell$}}\;\label{op:SVDComp}
		}
	}	
	\vspace{1em}
	\caption{Background Subtraction using adaptive SVD.}
	\label{alg:AdaptiveSVD}
\end{algorithm}
%\vspace{-1em}

\noindent
The algorithm uses the following parameters:
\begin{itemize}	
\item $\ell$: Parameter used in \textit{SVDComp} for rank-$\ell$ approximation.
\item $\eta$: Parameter for setting up the maximal Frobenius norm as a
  multiple of the Frobenius norm of an average image. 
\item $\tau^*$: Threshold value for the slope of the singular values
  used in \textit{SVDAppend}. 
\item $\theta$: Threshold value depending on the pixel intensity range
  to discard noise in the foreground image. 
\item $\beta$: Number of frames put together to one block $B_k$ for
  \textit{SVDAppend}. 
\item $n^*$: Maximum number of columns of $U_k$. If $n^*$ is reached a
  re-initialization is triggered.
\end{itemize}

For the exposition in Alg.~\ref{alg:AdaptiveSVD} we use pseudo-code
with a MATLAB like syntax. Two further explanations are necessary,
however. First, we remark that \textit{SVDAppend} and \textit{SVDComp}
return the updated matrices $U$ and $\Sigma$ and the \emph{index} of
the thresholding singular value determined by $\tau^*$ as described in
Sec.~\ref{subsubsec:svdappend}.
Using the threshold value $\theta$, the foreground resulting from the
subtraction of the background from the input image gets
binarized. This binarization is used as mask on the input image to
gain the parts that are considered as foreground. $|B - J| > \theta$
checks elementwise whether $| B_{jk} - J_{jk} | > \theta$ and returns
a matrix consisting of the Boolean values of this operation.

\subsection{Relaxation of the small foreground assumption}
\label{subsec:relaxation}
\begin{figure*}[htb]
	\begin{subfigure}{.5\textwidth}
		\centering
		\includegraphics[width=.8\linewidth]{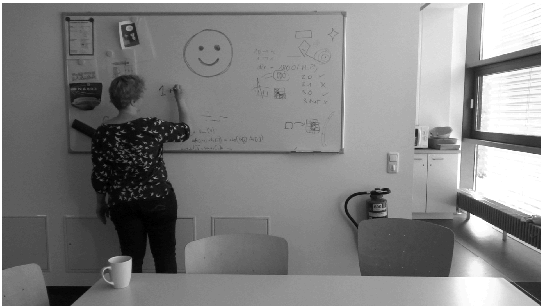}
		\caption{Original image.}
		\label{fig:stock_effect_1}
	\end{subfigure}%
	\begin{subfigure}{.5\textwidth}
		\centering
		\includegraphics[width=.8\linewidth]{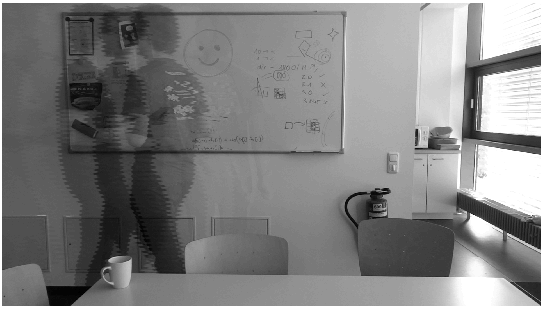}
		\caption{Orthogonal projection onto background subspace.}
		\label{fig:stock_effect_2}
	\end{subfigure}
	\begin{subfigure}{.5\textwidth}
		\centering
		\includegraphics[width=.8\linewidth, clip, trim=0 0 0 -20]{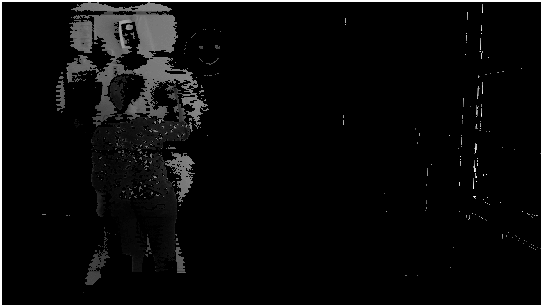}
		\caption{Foreground image.}
		\label{fig:stock_effect_3}
	\end{subfigure}%
	\begin{subfigure}{.5\textwidth}
		\centering
		\includegraphics[width=.8\linewidth, clip, trim=0 0 0 -20]{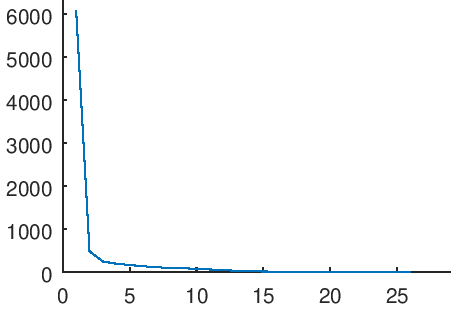}
		\caption{Magnitude of the singular values plotted over the position on the diagonal of $\Sigma$.}
		\label{fig:stock_effect_4}
	\end{subfigure}
	\caption{Example of artifacts due to a big foreground object
          that was added to the background. The foreground object in
          the original image (a) triggers singular vectors containing
          foreground objects falsely added to the background (b) in
          previous steps. These artifacts can thus be seen in the
          foreground image (c).}
	\label{fig:stock_effect}
\end{figure*}

A basic assumption of our background subtracting algorithm is that the
changes due to the foreground are small relative to the image
size. Nevertheless, this assumption is easily violated, e.g. by a
truck in traffic surveillance or generally by objects close to the
camera which can appear in singular vectors that should represent
background. This has two consequences. The first is that the
foreground object is not recognized as such, the second one leads to
ghosting effects because of the inner product as shown in
Fig.~\ref{fig:stock_effect}.

\noindent
The following modifications increase the robustness of our method
against these unwanted effect.

\subsubsection{Similarity Check}
\label{subsubsec:sim_check}
Big foreground objects can exceed the threshold level $\tau$ in
\textit{SVDAppend} and therefore are falsely included in the
background space.
% This happens, because the assumption of small foreground objects is
% violated.
% REDUNDANT
With the additional assumption that background effects have to be
stable over \emph{time}, frames with large moving objects can be
filtered out by utilizing the block appending property of the adaptive
SVD. There, a large \emph{moving} object causes significant
differences in a block of images which can be detected by calculating
the structural similarity of a block of new images. Wang et al. propose
in~\cite{SSIM} the normalized covariance of two images to capture the
structural similarity. This again can be written as the
inner product of normalized images, i.e.,
\[
  s(B_i, B_j) = \frac{1}{d-1}\sum_{l=1}^{d}
  \frac{B_{i,l} - \mu_{i}}{\sigma_{i}} \frac{B_{j,l} - \mu_{j}}{\sigma_{j}},
\]
with $B_i$ and $B_j$ being two vectorized images with $d$ pixels,
means $\mu_i$, $\mu_j$ and standard deviations $\sigma_i$ and
$\sigma_j$.
% Let $\hat{B}_i := \frac{B_i - \mu_{B_i}}{\sigma_{B_i}}$ and $\hat{B}_j
% := \frac{B_j - \mu_{B_j}}{\sigma_{B_j}}$ be the normalized images then
% $s(B_i, B_j) = \hat{B}_i^T \hat{B}_j$, which is the orthogonal
% projection of $\hat{B}_j$ onto $\hat{B}_i$.
Taking into account that the input images already become normalized in
our algorithm, see Sec.~\ref{subsubsec:normalization}, this boils down
to a inner product. 

Given is a temporally equally spaced and ordered block of images $B := \{B_1, B_2, ... , B_m\}$ 
and one frame $B_i$ with $i \in \{1, 2, ... , m\} =: M$. 
The structural similarity of frame $B_i$ regarding the block $B$
is the measure we search for. This can be calculated by
\[
\frac{1}{m-1}\sum_{j \in M\setminus\{i\}} s(B_i,B_j),
\]
i.e., the mean structural similarity of $B_i$ regarding $B$. For the relatively short time span of
one block it generally holds that $s(B_i, B_j) \geq s(B_i, B_k)$ with $i, j, k \in M$
and $i < j < k$, i.e., the structural similarity drops going further into
the future as motions in the images imply growing differences. 
This effect causes the mean structural similarity of the first or last frames of $B$
generally being lower than of the middle ones due to the higher mean time difference
to the other frames in the block.  

This bias can be avoided by calculating the mean
similarity regarding subsets of $B$. Let $\nu > 0$ be a fixed number of pairs to be considered
for the calculation of the mean similarity and $\Delta T \in
\mathbb{N}^+$ be the fixed cumulative time difference. Calculate the
mean similarity $\overline{s_i}$ of $B_i$ regarding to $B$ by selecting pairwise
distinct $\{j_1, j_2, ..., j_{\nu}\}$ from $M\setminus\{i\}$ with 
\[\sum_{l=1}^{\nu} |j_l - i| = \Delta T\quad \text{and}\quad \overline{s_i} = \frac{1}{\nu} \left(\sum_{l=1}^{\nu} s(B_i, B_{j_l})\right).\]
If $\overline{s_i}$ is smaller than the predefined similarity
threshold $\overline{s}$, frame $i$ is not considered for the
\textit{SVDAppend}.

\subsubsection{Periodic Updates}
\label{subsubsec:per_up}
Using the threshold $\tau$ speeds up the iterative process, but also
has a drawback: if the incoming images stay constant over a longer
period of time, the background should mostly represent the input
images and there should be high singular values associated to the
singular vectors describing it. Since input images that can be
explained well do not get appended anymore, this is, however, not the
case.
Another drawback is that outdated effects, like objects that stayed in
the focus for quite some time and then left again, have a higher
singular vectors than they should, as they are not relevant any more.
Therefore, it makes sense to periodically append images although they
are seen as irrelevant and do not surpass $\tau$. This also helps to
remove falsely added foreground objects much faster.

\subsubsection{Effects of the re-initialization strategy}
\label{subsubsec:re-initialization}

The re-initialization strategy (II) based on the background images
$\cols{U_k}{\hat{i}} (\cols{U_k}{\hat{i}} ^T B_i)$ as described in
Sec.~\ref{subsubsec:reinitialization} supports the removal of
incorrectly added foreground objects. When such an object, say $X$,
is gone from the scene, i.e., $B_i$ does not contain $X$ and
$\cols{U_k}{\hat{i}} (\cols{U_k}{\hat{i}} ^T B_i)$ does not contain it
either because a singular vector not containing $X$ approximates $B_i$
much better. As $X$ was added to the background, there must be at
least one column $j^*$ of $U_k$ containing $X$, i.e.,
$\cols{U_k}{j^*}^T X \gg 0$. As $\cols{U_k}{\hat{i}}
(\cols{U_k}{\hat{i}} ^T B_i)$ does not contain $X$,
$(\cols{U_k}{\hat{i}}^T B_i)_{j^*}$ must be close to zero as otherwise
the weighted addition of singular vectors $\cols{U_k}{\hat{i}}
(\cols{U_k}{\hat{i}}^T B_i)$ cancels $X$ out. The re-initialization is
thus based on images not containing $X$ and the new singular vectors also
do not contain leftovers of $X$ anymore.

Finally, the parameter $\eta$ modifies the size of the maximum
Frobenius norm used for normalization in re-initialization strategy
(III) from Sec.~\ref{subsubsec:normalization}. A smaller $\eta$
reduces the importance of the already determined singular vectors
spanning the background space and increases the impact of newly
appended images. If an object $X$ was falsely added, it gets removed
more quickly if current frames not containing $X$ have a higher
impact. A similar behavior like with re-initialization strategy (II)
can be achieved. The disadvantage is that the background model changes
quickly and does not capture long time effects that well. In the end,
it depends on the application which strategy performs better. 

\begin{figure*}[tbh]
	\begin{subfigure}{.5\textwidth}
		\centering
		\includegraphics[width=.9\linewidth]{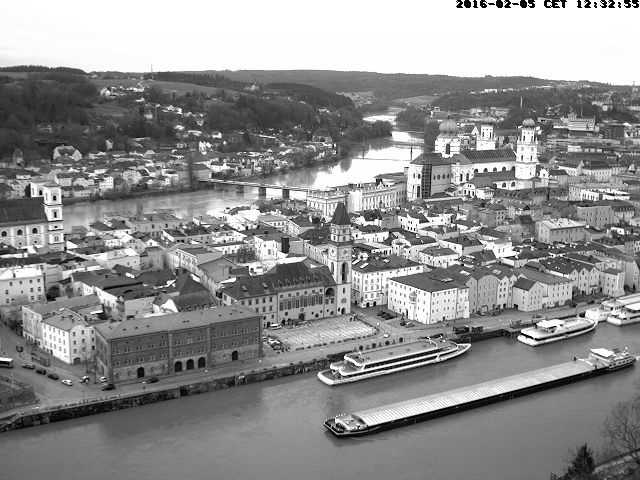}
		\caption{Original image.}
		\label{fig:passau_example:a}
	\end{subfigure}%
	\begin{subfigure}{.5\textwidth}
		\centering
		\includegraphics[width=.9\linewidth]{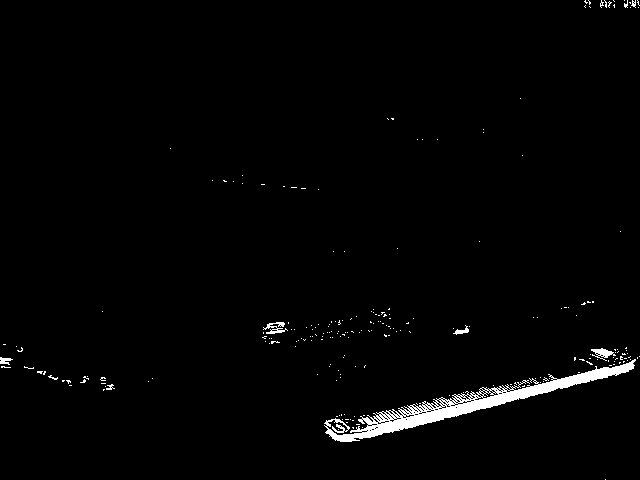}
		\caption{Foreground image.}
		\label{fig:passau_example:b}
	\end{subfigure}
	\caption{Example frame from a webcam video monitoring the city of Passau. In~\ref{fig:passau_example:a} the input image can be seen and in~\ref{fig:passau_example:b} the foreground image as a result of algorithm~\ref{alg:AdaptiveSVD}.}
	\label{fig:passau_example}
\end{figure*}

\section{Computational Results}
\label{sec:comp_results}
The evaluation of our algorithm is done based on an implementation in the \CC\ programming language using Armadillo~\cite{Armadillo} for linear Algebra computations.

\subsection{Default Parameter Setting}
\label{subsec:default_parameters}
Alg.~\ref{alg:AdaptiveSVD} depends on parameters that are still to be
specified. In the following, we will introduce a default parameter
setting that works well in many different applications. The parameters
could even be improved or optimized for a specific application using
ground truth data. Our aim here, however, is to show that the adaptive
SVD algorithm is a very generic one and applicable almost ``out of the
box'' for various situations. The chosen \emph{default parameters} are
as follows:
\begin{itemize}
	\item $\ell = 15$, %cut_back	
	\item $n^* = 30$, % re-initialization at n=n^*
	\item $\eta = 30$, % system_size
	\item $\tau^* = 0.05\cdot\rho$, with $\rho = \frac{||A||_F}{\sqrt{n}}\sqrt{\eta}$ of the initialization matrix $A$,  % imp_vecs_threshold
	\item $\beta = 6$, $\nu = 3$, $\Delta T = 6$, $\overline{s} = 0.97$,  % block_size
	\item $\theta = 1.0$.  % output threshold
\end{itemize} 
The parameter $\ell$ determines how many singular values and
corresponding singular vectors are kept after
re-initialization. Setting $\ell$ too low can cause a loss of
background information. In our examples, $15$ turned out to be
sufficient not to lose information. The re-initialization is triggered
when $n^*$ relevant singular values have been accumulated. Choosing
that parameter too big reduces the performance, as the floating point
operations per \textit{SVDAppend} step depend cubically on the number
of singular vectors and linearly on the number of singular vectors
times the number of pixels, see Sec.~\ref{subsec:IterSVD}.
The system size $\eta$ controls the impact of newly appended frames,
and a large value of $\eta$ favors a stable background.
The threshold value $\tau^*$ for the discrete slope of singular values
in the \textit{SVDAppend} step depends on the data. The heuristic factor $0.05$
proved to be effective to indicate that the curve of the singular
values flattens out.
The block size $\beta$ and the corresponding $\nu$ and $\Delta T$
depend on the frame rate of the input. The choice is such that it does
not delay the update of the background space too much, which would be
the effect of a large block size. Keeping it relatively small, we are
able to evaluate the input regarding similarity and stable
effects. Due to the normalization of the input images to zero mean and
standard deviation one, the similarity threshold $\overline{s}$ and
the binarization threshold $\theta$ are stable against different input
types.

\subsection{Small Foreground Objects}

The first example video for a \emph{qualitative} evaluation is from a
webcam monitoring the city of Passau, Germany, from above. The
foreground objects, e.g. cars, pedestrians, boats, are small or even
very small. The frame rate of 2 frames per minute is relatively low
and the image size is $640 \times 480\,$px. This situation allows for
a straightforward application of the basic adaptive SVD algorithm
without similarity check and regular updates. The remaining parameters
are as in the default setting of
Sec.~\ref{subsec:default_parameters}.

\begin{figure}[bht]
	\centering
	\includegraphics[width=.95\linewidth]{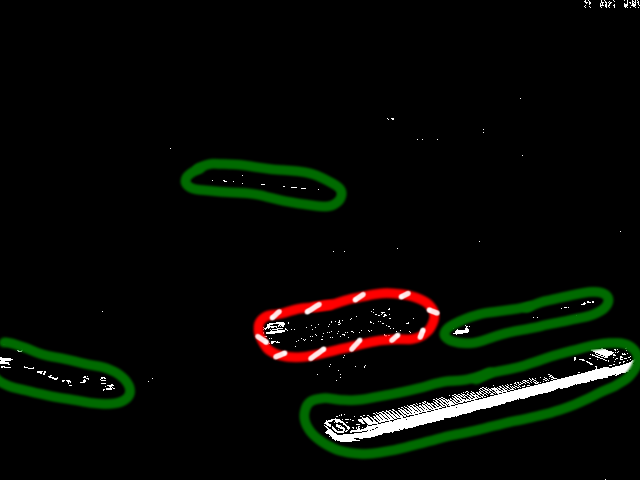}
	\caption{Plot marking the true detections in the foreground
          image of Fig.~\ref{fig:passau_example:b} by green circles
          and incorrect detections by red circles with white stripes.}
	\label{fig:passau_example_marked}
\end{figure}

In Fig.~\ref{fig:passau_example} an example frame\footnote{\label{note1}The complete sample videos can be downloaded following \url{https://www.forwiss.uni-passau.de/en/media_and_data/}.} and the according
foreground image from the webcam video is shown. The moving boat in
the foreground, the cars in the lower left and right corners, and even the cars
on the bridge in the background are detected well. 
%There is someclutter due to reflections on the river and from the tied (and onlyslightly moving) boat in the foreground. 
Small illumination changes and reparking vehicles
lead to incorrect detections on the square in the front. Fig.~\ref{fig:passau_example_marked} 
depicts these regions.

\subsection{Handling of Big Foreground Objects}
\label{subsec:big_fg_obj}

\begin{figure*}[htb]
	\begin{subfigure}{.5\textwidth}
		\centering
		\includegraphics[width=.8\linewidth]{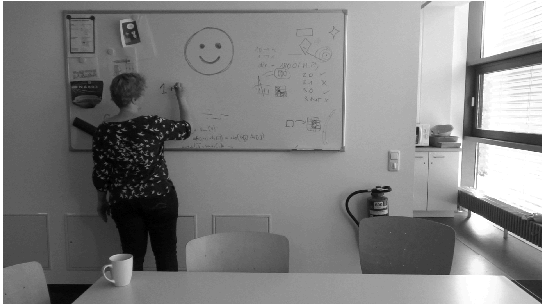}
		\caption{Original image.}
	\end{subfigure}%
	\begin{subfigure}{.5\textwidth}
		\centering
		\includegraphics[width=.8\linewidth]{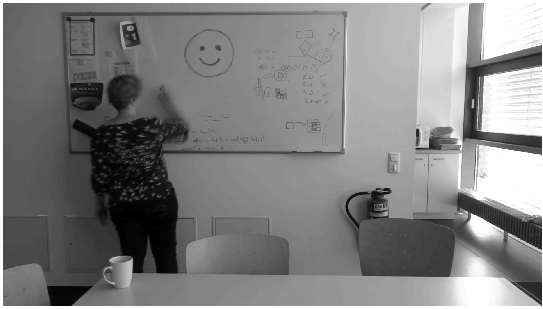}
		\caption{Orthogonal projection onto background subspace.}
	\end{subfigure}
	\begin{subfigure}{.5\textwidth}
		\centering
		\includegraphics[width=.8\linewidth]{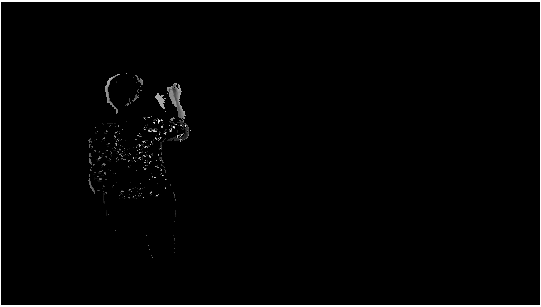}
		\caption{Foreground image.}
	\end{subfigure}%
	\begin{subfigure}{.5\textwidth}
		\centering
		\includegraphics[width=.8\linewidth]{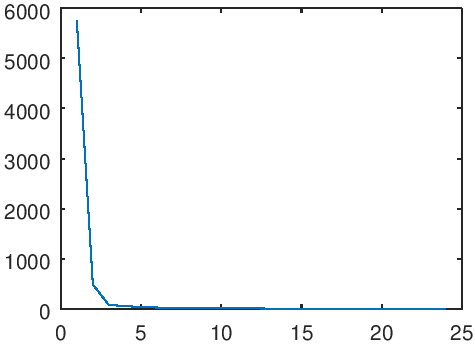}
		\caption{Magnitude of the singular values plotted over the position on the diagonal of $\Sigma$.}
	\end{subfigure}
	\caption{The same scene as in Fig.~\ref{fig:stock_effect}. The
          artifacts due to big foreground objects are reduced by
          similarity checks and regular updates. The current
          foreground object gets added to the background only after
          being stationary for a series of frames.} 
	\label{fig:stock_effect_removed}
\end{figure*}

Fig.~\ref{fig:stock_effect} is a frame from an example video\footnoteref{note1} including
the projection onto the background space, the computed foreground
image, and the distribution of the singular values. To illustrate the
improvements due to similarity checks and periodic updates, the same
frame is depicted in Fig.~\ref{fig:stock_effect_removed} where the
extended version of our algorithm is applied. The artifacts due to big
foreground objects that were added to the background in previous
frames, are not visible anymore. The person in the image still gets
added to the background, but only after being stationary for some
frames.

\subsection{Execution Time}
\label{subsec:execution_time}

The performance of our implementation is evaluated based on an
Intel\textregistered\ Core\texttrademark\ i7-4790 CPU @
\SI{3.60}{\hertz} $\times\ 8$. The example video from
Sec.~\ref{subsec:big_fg_obj} has a resolution of $1920 \times
1080\,$px with $25\,$fps. For the application of our algorithm on the
example video, the parameters are set as shown in
Sec.~\ref{subsec:default_parameters}. 

As the video data was recorded with $25\,$fps, there is no need to
consider every frame for a background update, because the background
is assumed to be constant over a series of frames and can only be
detected considering a series of frames. Therefore, only every second
frame is considered for a background update, while a background
subtraction using the current singular vectors is performed  on every
frame. Our implementation with the settings from
Sec.~\ref{subsec:default_parameters} handles this example video with $8\,$fps.

For surveillance applications it is important that the background
subtraction is applicable in real time for which $8\,$fps are too
slow. One approach would be to reduce the resolution. The effects of
that will be discussed in the following section. Leaving the
resolution unchanged, the parameters have to be adapted. Setting
$\ell=10$ and $n^*=25$ significantly reduces the number of background
effects that can be captured, but turns out to be still sufficient for
this particular scene. The number of images considered for background
updates can be reduced as well. Downsampling the frames by averaging over a
window size of $8$ and setting $\ell=10$ and $n^*=25$ leads to a processing rate of
$25\,$fps which is \emph{real time}.

In Sec.~\ref{subsec:IterSVD} we pointed out that the number of
floating point operations for an update step depends linearly on the
number of pixels $d$ when using Householder reflections. A
re-initialization step is computationally even cheaper, because only
Householder vectors have to be updated. The following execution time
measurements underline the theoretical considerations. 
Our example video is resized several times, 900 images are appended,
and re-initialization is performed when $n^*$ singular vectors are
reached. Tab.~\ref{tab:linear_time} shows the summed up time for the
append and re-initialization steps during iteration for the given
image sizes. The number $d$ of pixels equals $\num[group-separator={,}]{2073600} = 1920\cdot1080$.

\begin{table} [htb]
	%	\vskip -4mm
	\begin{center}
		%\resizebox{8.7cm}{!}{
		\begin{tabular}{ | c | c | c | c | c | }
			\hline
			\#Pixels & Append & Factor & Re-init. & Factor\\ \hline
			$d$ & \SI{69.30}{\second} & 1.90 & \SI{16.46}{\second} & 1.37 \\
			$d/2$ & \SI{34.28}{\second} & 1.88 & \SI{8.25}{\second} & 1.38 \\
			$d/4$ & \SI{16.67}{\second} & 1.83 & \SI{3.79}{\second} & 1.26 \\
			$d/8$ & \SI{6.72}{\second} & 1.47 & \SI{1.68}{\second} & 1.12 \\
			$d/16$ & \SI{2.28}{\second} & 1 & \SI{0.75}{\second} & 1 \\
			\hline
		\end{tabular}%}
	\end{center}
	\caption{Execution time for performing a SVD update iteratively on 900 frames for different image sizes and $d = 2073600 = 1920\cdot 1080$. \label{tab:linear_time}}
\end{table}

The factors $t_{d/i} / (t_{d/16} \cdot \frac{16}{i})$ with $i \in
\{1,2,4,8,16\}$ and total append or re-initialization times $t_{d/i}$
are shown in Tab.~\ref{tab:linear_time} for image sizes $d/i$. These
factors should be constant for increasing image sizes due to the
linear dependency. Still, the factors keep increasing, but even less
than a logarithmic order. This additional increase in execution time
can be explained due to the growing amount of memory that has to be
managed and caching becomes less efficient as with small images.

\subsection{Evaluation on Benchmark Datasets}
\label{subsec:benchmark}
\begin{figure}[htb]
	\centering
	\includegraphics[width=.8\linewidth]{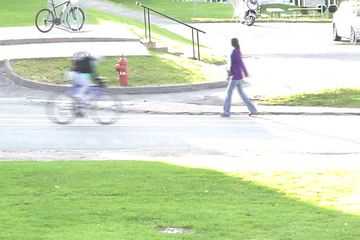}
	\caption{Example frame from the \textit{pedestrians} video of the CDnet database.}
	\label{fig:pedestrians_frame}
\end{figure}

The quantitative evaluation is performed on example videos from the
background subtraction benchmark data set CDnet
2014~\cite{CDnet14}. The first one is the \textit{pedestrians} video
belonging to the \textit{baseline} category. It contains 1099 frames
($360 \times 240$ px) of people walking and cycling in the public. An
example frame can be seen in Fig.~\ref{fig:pedestrians_frame}. 

For the frames 300 trough 1099 binary ground truth annotations exist
that distinguish between foreground and background. From the first 299 frames, 15 frames are equidistantly sub-sampled and taken for the initial matrix $M$. Thereafter, Alg.~\ref{alg:AdaptiveSVD} is executed on all frames from 300 through 1099. Instead of applying the binary mask in line 18 of algorithm~\ref{alg:AdaptiveSVD} onto the input image, the mask itself is the output to achieve binary images.

\begin{table*} [htb]
	%	\vskip -4mm
	\begin{center}
		%\resizebox{8.7cm}{!}{
		\begin{tabular}{ | c | c | c | c | c | c | c | c | }
			\hline
			& Recall & Specificity & FPR & FNR & PBC & Precision & F-Measure \\ \hline
			\textit{default} & 0.869 & 1.000 & 0.000 & 0.131 & 0.158 &	0.967 &	0.915 \\
			\textit{morph} & 0.936 & 1.000 & 0.000 & 0.063 & 0.088 & 0.973 & 0.954 \\
			\hline
		\end{tabular}%}
	\end{center}
	\vskip -4mm
	\caption{Evaluation of the \textit{pedestrians} scene of the
          CDnet database with the benchmark evaluation metrics
          including FPR (False Positive Rate), FNR (False Negative
          Rate), PBC (Percentage of Wrong
          Classifications).\label{tab:cdnet_eval}} 
\end{table*}

With the default parameter setting of
Sec.~\ref{subsec:default_parameters} a pixelwise \emph{precision} of
$0.958$ and an \emph{F-measure} of $0.919$ are achieved with a
performance of 843 fps.
The thresholding leading to the binary mask is sensitive to the
contrast of the foreground relative to the background. If it is low,
foreground pixels are not detected properly. To avoid missing pixels
within foreground objects, the morphological close operation is
performed with a circular kernel. Moreover, a fixed minimal size of
foreground objects can be assumed reducing the number of false
positives. These two optimizations lead to a precision of $0.968$ and
an F-measure of $0.958$ at $684$ fps. The complete evaluation measures
can be seen in Tab.~\ref{tab:cdnet_eval}. \textit{Default} represents
the default parameter setting and \textit{morph} the version with the
additional optimizations. In the following, the morphological
postprocessing is always included. 

Our method delivers a state of the art performance for unsupervised
methods. The best unsupervised method, IUTIS-5~\cite{IUTIS-5} on the
benchmark site could achieve a precision of $0.955$ and an F-measure of
$0.969$. It is based on genetic programming combining other state of
the art algorithms. The execution time is not given, but naturally
higher than the execution time of the slowest algorithm used, assuming
perfectly parallel execution. We introduced domain knowledge only in
the morphological optimizations. Otherwise, there is no specific
change towards the test scene.
Even more domain knowledge is used in supervised learning techniques
as object shapes are trained and irrelevant movements in the
background are excluded due to labeling. They are able to outperform
our approach regarding the evaluation measures. An overall average
precision and F-measure of more than $0.98$ is achieved. The benchmark
site disclaims, nevertheless, that the supervised methods may have
been trained on evaluation data as ground truth annotations are only
available for evaluation data. 

The positive effect of a block-wise appending of the data with a
similarity check and regular updates as shown above also applies here:
our adaptive SVD algorithm on the given \textit{pedestrians} video
from the benchmark site without using the similarity checks and
regular updates only leads to a precision of $0.933$ and an F-measure of
$0.931$.

The performance of our algorithm on more example videos from the CDnet
data set is listed in Tab.~\ref{tab:cdnet_eval_more}. The
\textit{park} video is recorded with a thermal camera, the
\textit{tram} and \textit{turnpike} videos with a low frame rate, and
the \textit{blizzard} video while snow is falling.
For \textit{highway} and \textit{park} the best unsupervised method is
IUTIS-5 and for \textit{tram}, \textit{turnpike}, \textit{blizzard},
and \textit{streetLight} that is
SemanticBGS~\cite{SemanticBGS}. SemanticBGS combines IUTIS-5 and a
semantic segmentation deep neural network and the execution time is
given with 7 fps for $473 \times 473\,$px images based on a NVIDIA
GeForce GTX Titan X GPU. 

\begin{table} [htb]
	%	\vskip -4mm
	\begin{center}
		%\resizebox{8.7cm}{!}{
		\begin{tabular}{ | c | c | c | c | c | }
			\hline
			Video & Prec & F-Meas & Prec* & F-Meas* \\ \hline
			\textit{highway} & 0.901 & 0.816 & 0.935 & 0.954 \\
			\textit{park} & 0.841 & 0.701 & 0.776 & 0.765 \\
			\textit{tram} & 0.957 & 0.812 & 0.838 & 0.886  \\
			\textit{turnpike} & 0.962 & 0.860 & 0.980 & 0.881  \\
			\textit{blizzard} & 0.919 & 0.854 & 0.939 & 0.845 \\
			\textit{streetLight} & 0.992 & 0.982 & 0.984 & 0.983 \\
			\hline
		\end{tabular}%}
	\end{center}
	\vskip -4mm
	\caption{Evaluation of the adaptive SVD algorithm on example
          videos from the CDnet data set using precision and
          F-measure. Prec$^*$ and F-Meas$^*$ give the precision and
          F-measure of the best unsupervised method of the benchmark
          regarding to the test video.\label{tab:cdnet_eval_more}} 
\end{table}

Besides the \textit{park} video, the content is mostly vehicles
driving by. The performance of our algorithm clearly drops whenever
the initialization image set contains a lot of foreground objects like
in the \textit{highway} video, where the street is never empty.
Moreover, a foreground object turns into background when it stops
moving which is even a feature of our algorithm. This, however, causes
problems in a lot of the benchmark videos of the CDnet benchmark with
vehicles stopping at traffic lights, like in the \textit{tram} video,
or people stopping and starting to move again.
There is a category of videos with intermittent object motion in the
CDnet data set. Our algorithm performs with an average precision of
$0.752$ and an F-measure of $0.385$ whereas SemanticBGS reaches an
average precision of $0.915$ and an F-measure of $0.788$.
The precision of our algorithm tends to be higher than the F-measure,
as it detects motion very well and therefore is certain that if there
is movement, it is foreground, but often foreground is not detected
due to a lack of motion. To delay the addition of a static object to
the background, it is possible to reduce the regular updates, for
example. But as this feature regulates the adaption of the background
model to a change in the background, this only enhances the
performance for very stable scenes. In the \textit{streetLight} video
no regular update was performed in contrast to the other
videos. Including regular updates, the precision is $0.959$ and the
F-measure $0.622$ due to cars stopping at traffic lights. The only
domain knowledge we introduce is the postprocessing via morphological
operations. Otherwise, the algorithm has no knowledge about the kind
of background it models. Therefore, not only vehicles or people are
detected as foreground, but also movement of trees or the reflection
of the light of the vehicles on the ground, which is negative for the
performance regarding the CDnet benchmark.

\section{Conclusions}
\label{sec:conclusions}
We utilized the iterative calculation of a Singular Value
Decomposition to model a common subspace of a series of frames which
is assumed to represent the background of the frames. An algorithm,
the \emph{adaptive SVD} was developed and applied for background
subtraction in image processing. The assumption that the foreground
has to be small objects was considered in more detail and relaxed by
extensions of the algorithm. In an extensive evaluation, the
capabilities of our algorithm were shown qualitatively and
quantitatively using example videos and benchmark results. Compared to
state of the art unsupervised methods we obtain competitive
performance with even superior execution time. Even high definition
videos can be processed in real time. 

The evaluation also showed that, if an application to a domain such as
video surveillance is intended, our algorithm would need to be
extended to also consider semantic information. Therefore, it can only
be seen as a preprocessing step, e.g. reducing the search space for
classification algorithms. In future work we aim to evaluate the
benefit of using our algorithm in preprocessing of an object
classifier. Moreover, we will address the issue of foreground objects
turning into background after being static for some time which is
desirable in some cases and erroneous in others. A first approach is
to use tracking, because objects do not disappear without any
movement. In the end, there is also some parallelization ability in
our algorithm separating the projection onto the background of
incoming images from the update of the background model. Further
performance improvements will be investigated.

\begin{acknowledgement}
Our work results from the project DeCoInt$^2$, supported by the German Research Foundation (DFG) within the priority program SPP 1835: "Kooperativ interagierende Automobile", grant numbers DO~1186/1-1, FU~1005/1-1, and SI~674/11-1.
\end{acknowledgement}

\bibliographystyle{spmpsci}      % mathematics and physical sciences
\bibliography{bibliography}   % name your BibTeX data base

\end{document}